\documentclass{article}

\usepackage{microtype}
\usepackage{graphicx}
\usepackage{subfigure}
\usepackage{booktabs} 

\usepackage{hyperref}

\usepackage[accepted]{icml2024}

\usepackage{amsmath}
\usepackage{amssymb}
\usepackage{mathtools}
\usepackage{amsthm}

\usepackage[capitalize,noabbrev]{cleveref}

\theoremstyle{plain}

\theoremstyle{definition}

\theoremstyle{remark}

\usepackage{cuted}

\usepackage[textsize=tiny]{todonotes}

\usepackage[textsize=tiny]{todonotes}

\usepackage{makecell}

\usepackage{mathtools}
\newcommand{\norm}[1]{\| #1 \|}
\newcommand{\bignorm}[1]{\Bigl \| #1 \Bigr \| }

\usepackage{multirow,tabularx}
\usepackage{adjustbox}

\usepackage{wrapfig}

\usepackage[title,toc,titletoc,page]{appendix} 
\icmltitlerunning{Leveraging Topological Guidance for Improved Knowledge Distillation}

\begin{document}

\twocolumn[
\icmltitle{Leveraging Topological Guidance for Improved Knowledge Distillation}



\icmlsetsymbol{equal}{*}

\begin{icmlauthorlist}
\icmlauthor{Eun Som Jeon}{yyy}
\icmlauthor{Rahul Khurana}{yyy}
\icmlauthor{Aishani Pathak}{yyy}
\icmlauthor{Pavan Turaga}{yyy}
\end{icmlauthorlist}

\icmlaffiliation{yyy}{Geometric Media Lab, Arizona State University, Tempe, AZ 85281, USA}

\icmlcorrespondingauthor{Eun Som Jeon}{ejeon6@asu.edu}

\icmlkeywords{Machine Learning, ICML}

\vskip 0.3in
]



\printAffiliationsAndNotice{}  






\begin{abstract}
Deep learning has shown its efficacy in extracting useful features to solve various computer vision tasks. However, when the structure of the data is complex and noisy, capturing effective information to improve performance is very difficult. To this end, topological data analysis (TDA) has been utilized to derive useful representations that can contribute to improving performance and robustness against perturbations. Despite its effectiveness, the requirements for large computational resources and significant time consumption in extracting topological features through TDA are critical problems when implementing it on small devices. To address this issue, we propose a framework called Topological Guidance-based Knowledge Distillation (TGD), which uses topological features in knowledge distillation (KD) for image classification tasks. We utilize KD to train a superior lightweight model and provide topological features with multiple teachers simultaneously. We introduce a mechanism for integrating features from different teachers and reducing the knowledge gap between teachers and the student, which aids in improving performance. We demonstrate the effectiveness of our approach through diverse empirical evaluations.
\end{abstract}

\section{Introduction} \label{sec:intro}
In recent years, deep learning has been widely deployed into various applications, such as image recognition \cite{xie2020self, he2019bag}, activity recognition \cite{zheng2016exploiting, wang2016human}, semantic segmentation \cite{minaee2021image}, and so on. 
Deep learning is proficient in extracting features and performing various computer vision tasks. However, it has challenges in grasping useful features from the complex structure of the data, which limits further advancements \cite{najafabadi2015deep}.

To address these issues, topological data analysis (TDA) has emerged as a solution, which is excellent at analyzing the topology of data to apprehend its arrangement \cite{adams2017persistence, WANG2021109324}. Since TDA reveals patterns that may not be extracted or magnified through traditional statistical methods, many research endeavors aim to adopt these attributes of TDA to enhance the efficacy of deep learning. Specifically, TDA is excellent in capturing inherent and invariant features, which are robust to noise and perturbation \cite{adams2017persistence,seversky2016time}.
TDA characterizes the shape of complex data, using the persistence of connected components and high-dimensional holes by the persistent homology (PH) algorithm. This persistence information can be represented as a persistence image (PI).
To utilize TDA in fusion of machine learning, PI has been widely used since it can be easily transformed and treated as a general image \cite{edelsbrunner2022computational}. 
Despite various benefits of TDA, significant computational resources and time is required for TDA feature computation. Many applications have explored the use of TDA features with machine learning \cite{Munch2017}, however in most cases, simple fusion methods do not result in compact models. Som \emph{et al.} \cite{som2020pi} introduced PI-net to solve this problem, however the burden of increased network size cannot be alleviated even at test-time.

\begin{figure}[t]
\begin{center}
\includegraphics[width = 0.45\textwidth]{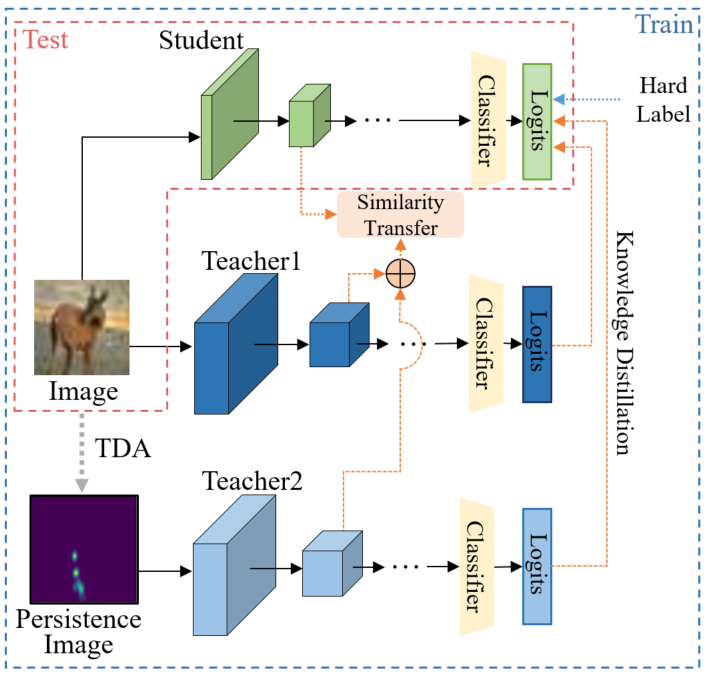}
\centering
\vskip -0.1in
\caption{An overview of Topological Guidance based Knowledge Distillation (TGD). Two teachers are trained with different representations from the raw image and persistence image data, respectively. A student utilizes the original image data alone.}
\label{figure:framework}
\end{center}
\vskip -0.15in
\end{figure}

Knowledge distillation (KD) has been addressed as a promising approach that leverages a power of a teacher (large model) to generate a student (small model) \cite{hinton2015distilling}. KD has further benefits in improving generalizability of a student model. In KD, a variety of strategies can be adopted to generate a compact model. For instance, not only one teacher model but multiple teachers can be utilized to transfer more diverse and strong knowledge to a student model \cite{gou2021knowledge}. An approach that involves utilizing two teachers can be adopted to leverage the power of topological knowledge. In detail, two teachers are trained -- one on the original data and the other one on the PI -- both of which are leveraged to generate a student model. This strategy has proven beneficial in time-series data analysis \cite{ejasilomar}. However, sufficient research has not been conducted on the effectiveness of such methods in KD based image analysis leveraging topological features. Additionally, when the statistical characteristics of knowledge from the two models are significantly different, there are considerable challenges and performance degradation in combining and utilizing the two sets of information \cite{zhu2021student, tan2018multilingual, gou2021knowledge}.

In this paper, we propose a framework, Topological Guidance based Knowledge Distillation (TGD), using topological knowledge in distillation for image classification task. We devise a strategy to integrate knowledge from two teachers trained with different modalities: raw image data and persistence images. 
An overview of the TGD is shown in Figure \ref{figure:framework}.
Firstly, PI is extracted from the raw image data through TDA. The extracted PI is then used to train a teacher model. Secondly, two teacher models are employed to provide useful information to train a student model. Logits and features from intermediate layers are utilized. 
When features of intermediate layers are transferred, similarity maps are utilized, facilitating the integration of information with different characteristics into a single entity. Additionally, we adopt an annealing strategy that reduces knowledge gap between teachers and students while preserving the weights that the student model needs to possess for its task \cite{ejasilomar}.
Finally, a student model is distilled, which uses solely the raw image data in test-time.

The contributions of this paper are as follows:
\begin{itemize}
\item We introduce a novel framework in knowledge distillation, using topological knowledge to generate a compact model for image classification tasks.
\item We devise a technique to integrate features from intermediate layers of teachers and a strategy to reduce the knowledge gap between teachers and student.
\item We demonstrate the effectiveness of leveraging topological features in KD empirically with various evaluations such as various combinations of teachers and students and feature visualizations.
\end{itemize}

\vskip -0.1in

Our main goal is not to outperform all the latest methods in vision, but to explore how topological guidance can be utilized in KD to improve the performance and to investigate the behavior of the distilled model along with empirical testing on image analysis.

\medskip


\section{Background} \label{sec:backg}
\subsection{Topological Feature}

TDA algorithms are applied to the data to extract topological features, which are robust to noises or perturbations and encodes the shape of complex data
\cite{adams2017persistence, WANG2021109324}. 
Persistent homology is a fundamental tool in TDA that helps in understanding the shape and structure of data, which involves constructing a filtration \cite{edelsbrunner2022computational}, typically based on a distance function and tracking variations of $n$-dimensional holes represented by assortments of points, edges, and triangles through a dynamic thresholding process called filtration.
In filtration, the appearance and disappearance of these holes are described in the persistence feature, summarized in a persistence diagram (PD), which records the birth and death times as $x$ and $y$ coordinates of planar scatter points \cite{adams2017persistence, edelsbrunner2022computational}.

\begin{figure}[htb!]

\begin{center}
\includegraphics[width = 0.48\textwidth]{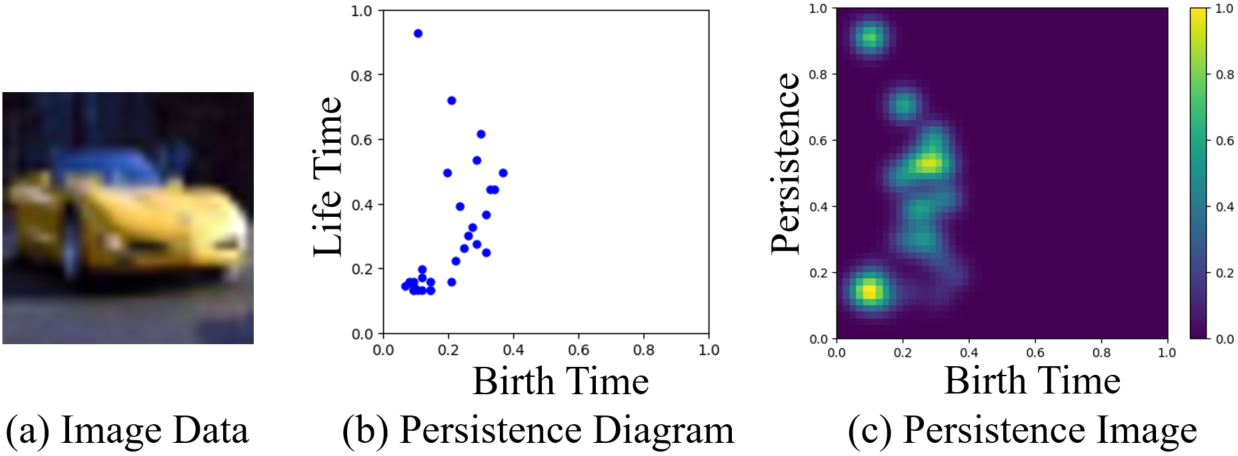}
\centering
\vskip -0.1in
\caption{PD and its corresponding PI. Lifetime points in PD appears bright colors in PI.}
\label{figure:PD_PI}
\end{center}
\vskip -0.1in
\end{figure}

Since PDs can have a high dimensionality with complex structures or a large number of points that can vary, using PDs directly in machine learning is challenging. 
To address this problem, persistence image (PI) has been widely used, which is one of the ways to encode geometric information via the lifespan of homological structures present in the data. This representation can be easily integrated into machine learning
\cite{barnes2021comparative, edelsbrunner2022computational}.
Specifically, the points within the PD are projected onto a two-dimensional grid $\rho: \mathbb{R} \rightarrow \mathbb{R}^2$. The grid points are then assigned values determined by a weighted sum of Gaussian smoothing, which is centered around the scattered points within the PD.
The grid is represented as a matrix and can be treated as regular image data called PI, as depicted in Figure \ref{figure:PD_PI}. This allows for the application of convolutional neural networks (CNNs) and machine learning algorithms, and offers a more manageable format for analysis and visualization.

Previous studies showed that topological features complement features of the raw data to achieve improved performance \cite{som2020pi}. 
However, additional processing for TDA and concatenation in networks increase computational time and resources at inference-time, which poses difficulty in implementing the process on small devices having limited computational resources and power.
To alleviate this issue, we propose a framework based on KD to infuse topological features into a small model that uses solely the raw image data at test-time.

\subsection{Knowledge Distillation}
\textbf{Conventional Knowledge Distillation.}
Knowledge distillation obtains a smaller model by utilizing the learned knowledge of a larger model, which was first introduced by Buciluǎ \emph{et al.} \cite{bucilua2006model} and explored more by Hinton \emph{et al.} \cite{hinton2015distilling}.
In KD, soft labels are utilized for knowledge transfer from a teacher to a student, which provide richer supervision signals and reduce overfitting. This also leads to better transferability of learned representations. The loss function of conventional KD for training a student is:
\begin{equation}
    \mathcal{L_{KD}} = \lambda\mathcal{L_{CE}} + (1- \lambda) \mathcal{L_{K}},
\end{equation}
where, $\mathcal{L_{CE}}$ is the standard cross entropy loss, $\mathcal{L_{K}}$ is KD loss, and $\lambda$ is a hyperparameter; $0 < \lambda < 1$. The difference between the output of the softmax layer for a student network and the ground-truth label is minimized by the cross-entropy loss:
\begin{equation}
    \mathcal{L_{CE}} = \mathcal{H}(softmax(l_{S}), y_g),
\end{equation}
where, $\mathcal{H(\cdot)}$ is a cross entropy loss function, $l_S$ is the logits of a student, and $y_g$ is a ground truth label. 
The gap between outputs of student and teacher are minimized by KL-divergence loss:
\begin{equation}\label{eq3}
    \mathcal{L_{K}} = \tau^{2}KL(z_{T}, z_{S}),
\end{equation}
where, $z_T = softmax(l_T/\tau)$ and $z_S = softmax(l_S/\tau)$ are softened outputs of a teacher and student, respectively, and $\tau$ is a hyperparameter; $\tau > 1$.
To obtain the best performance, we adopt early stopping for KD (ESKD) which improves the efficacy of KD \cite{cho2019efficacy}.

\textbf{Feature-based Knowledge Distillation.}
Features from intermediate layers of a network can be utilized in knowledge transfer \cite{gou2021knowledge, zagoruyko2016paying, tung2019similarity}. 
Zagoruyko \emph{et al.} \cite{zagoruyko2016paying} suggested activation-based attention transfer (AT), which is computed by a sum of squared attention mapping function, and calculating statistics across the channel dimension.
Tung \emph{et al.} \cite{tung2019similarity} introduced similarity-preserving knowledge distillation, matching similarity within a mini-batch of samples between a teacher and a student.
Since the size of a similarity map is determined by the size of a mini-batch, the size of the extracted similarity maps from the teacher and student is the same even if they generate different sizes of features.
In details, the similarity map $M \in \mathbb{R}^{b \times b}$ is obtained as follows:
\begin{equation}\label{eq_sp}
 M = F \cdot F^{\top}; F \in \mathbb{R}^{b \times chw},
\end{equation}
where $F$ is reshaped features from an intermediate layer of a model, $b$ is the size of a mini-batch, and $c$, $h$, and $w$ are the number of channels, height, and width of the output, respectively.
These feature transfer methods are popularly used; however, these are to match knowledge with similar characteristics in a uni-modal manner.

\textbf{Utilizing Multiple Teachers.}
Not only one teacher, but multi-teacher distillation has been widely utilized to provide more diverse knowledge in training process \cite{reich2020ensemble, liu2020adaptive,  gou2021knowledge}.
In some cases, the data utilized for training a student cannot be used during testing. Also, teachers trained with different modalities or representations can be utilized in distillation. Thoker and Gall \cite{thoker2019cross} train a student with paired samples from two modalities for action recognition. Jeon \emph{et al.} \cite{ejasilomar} explored to train a student model with two teachers trained with different representations for wearable sensor data analysis. With this insight, we develop a framework and explore to utilize topological features involving two teachers for image data analysis.

\section{Proposed Method} \label{sec:proposed}

In this section, we describe our proposed method -- TGD. Firstly, PI is extracted from an image through TDA. The extracted PI is utilized to train a teacher model. Secondly, we train a student model in KD with two teachers trained on different representations, the raw image data and PI. Then, to provide more useful knowledge to a student, features from teachers are integrated by considering correlations of each teacher's features. To reduce knowledge gap between teachers and student, an annealing strategy is applied. 

\subsection{Persistence Image Extraction}
To compute PIs, Scikit-TDA python library \cite{scikittda2019} and the Ripser package are used for generating PDs, as explained in Som {\em et. al.} \cite{som2020pi}. 
Firstly, image data is normalized in range in $[0, 1]$.
To compute level-set filtration PDs, image data is reshaped to row- and column-wise signals, considering different order of context which can extract different topological features \cite{barnes2021comparative}.
By filtration, PDs summarize the different peak and local minima intensities in the data.
Specifically, each channel of data is transformed and used to generate a PI, where the PI implies birth-time vs. lifetime information.
We utilize row- and column-wise transforms separately for creating images with channels of ($R_{r}$, $G_{r}$, $B_{r}$) and ($R_{c}$, $G_{c}$, $B_{c}$) to collect diverse knowledge, and all created PIs are concatenated in an image. Then, six channels ($P_{Rr}$, $P_{Gr}$, $P_{Br}$, $P_{Rc}$, $P_{Gc}$, $P_{Bc}$) of PI implying persistence knowledge is created. The total dimension size of one PI is $g\times g \times c$, where $g$ and $c$ are a constant value and the number of channels for a sample. 
The created PI is utilized to train a teacher model that acts as a pre-trained model to transfer topological features to a student model in KD process.

\subsection{Utilizing Multiple Teachers}

\textbf{Knowledge Transfer with Logits.}
Knowledge of logits from two teachers are transferred individually, thus additional process including concatenation
or hidden layers is not required. KD loss for logit knowledge of two teachers is explained as follows:
\begin{equation}
 \scalebox{0.99}{$
    \mathcal{L_{KD}}_l = \tau^{2} \left( \alpha KL(z_{T_1}, z_{S}) + (1-\alpha) KL(z_{T_2}, z_{S})\right),
$}
\end{equation}
where, $\alpha$ is a constant to balance the effects of different teachers, and $z_{T_1}$ and $z_{T_2}$ are softened outputs of teachers trained with the raw image data and PIs, respectively.

\begin{figure}[htb!]
\vskip -0.05in
\begin{center}
\includegraphics[width = 0.47\textwidth]{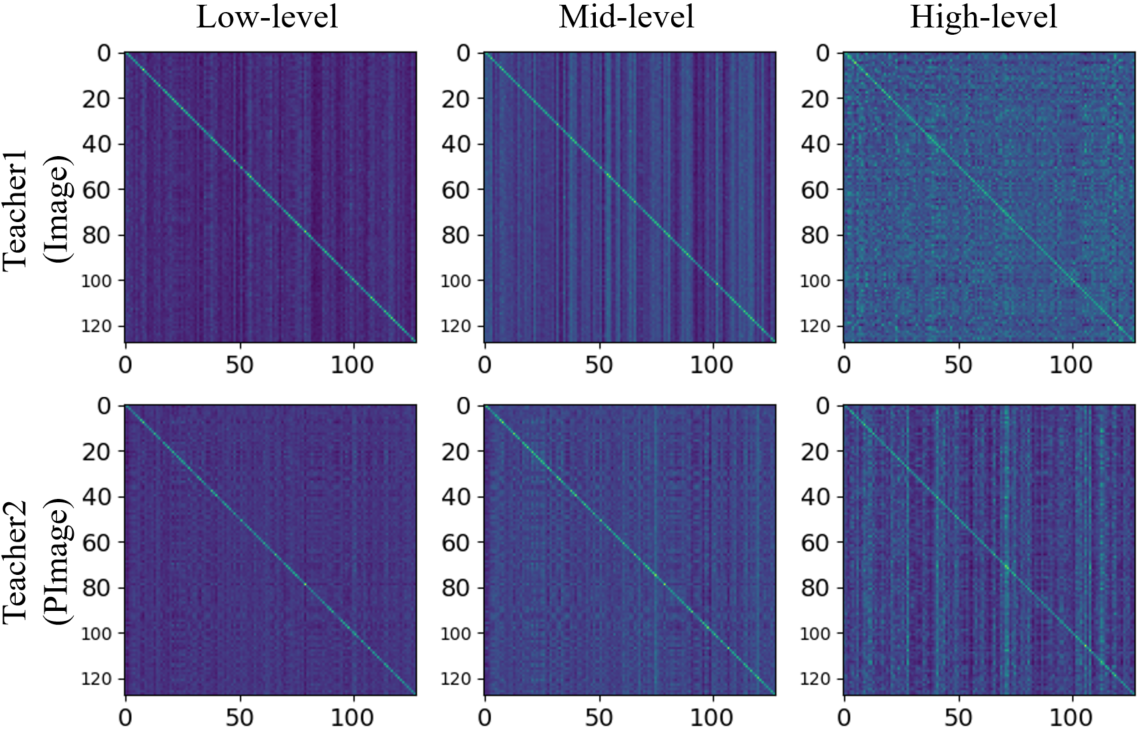}
\centering
\vskip -0.1in
\caption{An illustration of similarities for two teachers, trained with the raw image and persistence image, respectively.}
\label{figure:teacher_sim}
\end{center}
\vskip -0.15in
\end{figure}

\textbf{Knowledge Transfer with Intermediate Features.} To transfer sufficient knowledge from two teachers, we utilize features from intermediate layers additionally. Since PI and the raw image data have different statistical characteristics in semantic information, it is more effective to align and convey the information by using the correlation between samples rather than using spatial information. We utilize similarities, as explained in equation \ref{eq_sp}, to easily integrate information from two teachers and provide topological features to student in distillation.
Figure \ref{figure:teacher_sim} shows an example of similarity maps obtained from different intermediate layers of two WRN16-3 teachers. Note, Teacher1 and Teacher2 denote models learned with the raw image data and PI, respectively. High values represent high similarities. This implies similar patterns can be created when two samples belong to the same category. Since two models are trained with different representations, their highlighted patterns are different. We merge the maps from two teachers with weighted summation as follows:
\begin{equation}
    M^{(l)}_{T_m} = \alpha M^{(l^{T_1})}_{T_1} + (1-\alpha) M^{(l^{T_2})}_{T_2},
\end{equation}
where, $M^{(l)}_{T_m} \in \mathbb{R}^{b \times b}$ is the generated map from the similarity maps of two teachers $M_{T_1}$ and $M_{T_2}$ in a layer pair ($l^{T_1}$ and $l^{T_2}$). By merging the maps, the similarities include topological features which can complement the original features to improve the performance. The loss that encourages the student to mimic teachers is:
\begin{equation}
    \mathcal{L}_{m} = \frac{1}{b^2|L|}
    \sum_{(l, l^S) \in L} \biggl( \bignorm{ \widetilde{M^{(l)}_{T_m}} - \widetilde{M^{(l^S)}_{S}} }^{2}_{F} \biggr),
\end{equation}
where $\widetilde{M^{(l)}_{T_m}}$ and $\widetilde{M^{(l^S)}_{S}}$ are normalized map for a merged teacher and a student, $\norm{\cdot}_F$ is the Frobenius norm \cite{tung2019similarity}, and $L$ collects the layer pairs ($l$ and $l^S$). $l^{T_1}$, $l^{T_2}$, and $l^{S}$, can be selected with the same depth or the end of the same block of networks.
Since a student model uses the raw image data only, the gap between the merged features of teachers and the feature of the student can be generated, which makes degradation. To alleviate this problem, an annealing strategy \cite{ejasilomar} is used, which initializes the student model with weight values of a model trained from scratch.
The final loss function is as follows: 
\begin{equation}
    \mathcal{L}_\text{TGD} =  \lambda\mathcal{L_{CE}} + (1 - \lambda) \mathcal{L_{KD}}_{l} + \gamma\mathcal{L}_{m},
\end{equation}
where $\gamma$ is a hyperparameter.

\begin{table*}[htb!]
\renewcommand{\tabcolsep}{1.2mm} 
\centering
\vskip -0.1in
\caption{Details of teacher and student network architectures. ResNet \cite{he2016deep} and WideResNet \cite{zagoruyko2016wide} are denoted by ResNet (depth) and WRN (depth)-(channel multiplication), respectively. FLOPs and the number of trainable parameters correspond to one teacher model.}

\vskip 0.05in
\begin{center}
\begin{small}
\begin{sc}
\scalebox{0.92}{
\begin{tabular}{c |c |c |c |c |c |c |c| c| c}
\toprule
\centering
\multirow{2}{*}{DB}&  \multirow{2}{*}{Setup} &  \multirow{2}{*}{Compression type} & Teacher1 \& & \multirow{2}{*}{Student} & \multicolumn{2}{c|}{FLOPs} & \multicolumn{2}{c|}{\# of params} & Compression  \\ \cline{6-9}
& & & Teacher2 & & Teachers & Student & Teachers & Student & ratio  \\
 \midrule 
\multirow{4}{*}{\rotatebox[origin=c]{90}{\scriptsize{CIFAR-10}}}&(a) & Channel & WRN16-3 & WRN16-1 & 224.63M & 27.24M & 1.50M & 0.18M & 5.81$\%$ \\ 
&(b) & Depth & WRN28-1 & WRN16-1 & 56.07M & 27.24M & 0.37M & 0.18M & 24.32$\%$ \\ 
&(c) & Depth+Channel & WRN16-3 & WRN28-1 & 224.63M & 56.07M & 1.50M & 0.37M & 12.33$\%$ \\ 
&(d) & Different architecture & ResNet44 & WRN16-1 & 99.34M & 27.24M & 0.66M & 0.18M & 13.64$\%$ \\ 
\midrule

\multirow{4}{*}{\rotatebox[origin=c]{90}{\scriptsize{CINIC-10}}}&(a) & Channel & WRN16-3 & \multirow{4}{*}{WRN16-1} & 224.63M & \multirow{4}{*}{27.24M} & 1.50M & \multirow{4}{*}{0.18M} & 5.81$\%$ \\ 
&(b) & Depth & WRN28-1 & & 56.07M &  & 0.37M & & 24.32$\%$ \\ 
&(c$^a$) & Depth+Channel & WRN28-3 & & 480.98M & & 3.29M & & 2.74$\%$ \\ 
&(d) & Different architecture & ResNet44 & & 99.34M &  & 0.66M & & 13.64$\%$ \\ 
\bottomrule

\end{tabular}
}
\vskip -0.5in
\end{sc}
\end{small}
\end{center}
\label{table:info_settings}
\end{table*}

\begin{table*}[htb!]
\centering
\vskip -0.1in
\caption{Accuracy ($\%$) on CIFAR-10 with various knowledge distillation methods.}
\vskip 0.05in
\begin{center}
\begin{small}
\begin{sc}
\scalebox{0.93}{
\begin{tabular}{c |c c | c c c c c c c | c}

\toprule
\centering
\multirow{2}{*}{Setup} & \multicolumn{10}{c}{Method} \\ \cmidrule{2-11} 
& Teacher1 & Student & KD & AT & SP & RKD & VID & AFDS & Base & TGD \\ \midrule

\multirow{2}{*}{(a)} & 87.63 & 84.07 & 85.18 & 85.59 & 85.55 & 85.35 & 85.28 & \multirow{2}{*}{--} & 85.60 & \textbf{86.03}\\
& \scriptsize$\pm$0.09 & \scriptsize$\pm$0.08 & \scriptsize$\pm$0.14 & \scriptsize$\pm$0.08 & \scriptsize$\pm$0.05 &\scriptsize$\pm$0.06 & \scriptsize$\pm$0.19 & & \scriptsize$\pm$0.16 &\scriptsize$\pm$0.05 \\

\multirow{2}{*}{(b)} & 85.73 & 84.07 & 85.34 & 85.63 & 85.70 & 85.34 & 84.91 & 85.40 & 85.47 & \textbf{86.06} \\
& \scriptsize$\pm$0.06 & \scriptsize$\pm$0.08 & \scriptsize$\pm$0.15 & \scriptsize$\pm$0.06 & \scriptsize$\pm$0.07 &\scriptsize$\pm$0.10 & \scriptsize$\pm$0.25 &\scriptsize$\pm$0.19  &\scriptsize$\pm$0.17 &\scriptsize$\pm$0.14 \\

\multirow{2}{*}{(c)} & 87.63 & 85.73 & 86.38 & 86.63 & 86.44 & 86.16 & 86.35 & \multirow{2}{*}{--} & 86.86 & \textbf{87.12} \\
& \scriptsize$\pm$0.09 & \scriptsize$\pm$0.06 & \scriptsize$\pm$0.11 & \scriptsize$\pm$0.10 & \scriptsize$\pm$0.05 &\scriptsize$\pm$0.21 & \scriptsize$\pm$0.18 & & \scriptsize$\pm$0.11 &\scriptsize$\pm$0.06 \\

\multirow{2}{*}{(d)} & 86.15 & 84.07 & 85.36 & \textbf{85.91} & 84.69 & 85.43 &85.05 & 85.27 & 85.53 & 85.86 \\
& \scriptsize$\pm$0.11 & \scriptsize$\pm$0.08 & \scriptsize$\pm$0.10 & \scriptsize$\pm$0.09 & \scriptsize$\pm$0.12 &\scriptsize$\pm$0.08 & \scriptsize$\pm$0.09 &\scriptsize$\pm$0.17  & \scriptsize$\pm$0.08 &\scriptsize$\pm$0.04\\

\bottomrule

\end{tabular} }
\vskip -0.2in

\end{sc}
\end{small}
\end{center}

\label{table:results_CIFAR10}
\end{table*}


\begin{table*}[htb!] 
\centering
\vskip -0.1in
\caption{Accuracy ($\%$) on CINIC-10 with various knowledge distillation methods. TGD outperforms RKD \cite{park2019relational}. }
\vskip 0.05in
\begin{center}
\begin{small}
\begin{sc}
\scalebox{0.93}{
\begin{tabular}{c |c c | c c c c c c | c}

\toprule
\centering
\multirow{2}{*}{Setup} & \multicolumn{9}{c}{Method} \\ \cmidrule{2-10} 
& Teacher1 & Student & KD & AT & SP & VID & AFDS & Base & TGD \\ \midrule

\multirow{2}{*}{(a)} & 75.27 & \multirow{8}{*}{\shortstack{71.87 \\ {\scriptsize$\pm$0.09}}} & 74.20 & 74.32 & 74.25 & 74.31 & \multirow{2}{*}{--} & 74.43 & \textbf{74.66}\\
& \scriptsize$\pm$0.12 &  & \scriptsize$\pm$0.07 & \scriptsize$\pm$0.11 & \scriptsize$\pm$0.09 &\scriptsize$\pm$0.06 & &\scriptsize$\pm$0.26 &\scriptsize$\pm$0.04 \\

\multirow{2}{*}{(b)} & 73.41 &  & 74.57 & 74.51 & 74.81 & 73.75 & 74.45 & 74.71 & \textbf{74.88} \\
& \scriptsize$\pm$0.12 &   & \scriptsize$\pm$0.06 & \scriptsize$\pm$0.13 &\scriptsize$\pm$0.10 & \scriptsize$\pm$0.08 &\scriptsize$\pm$0.05 & \scriptsize$\pm$0.10 &\scriptsize$\pm$0.02 \\

\multirow{2}{*}{(c$^a$)} & 76.91 &  & 74.18 & 74.21 & 74.95 & 73.89 & \multirow{2}{*}{--} & 74.75 & \textbf{75.04} \\
& \scriptsize$\pm$0.03 &  & \scriptsize$\pm$0.06 & \scriptsize$\pm$0.10 & \scriptsize$\pm$0.16 &\scriptsize$\pm$0.16 & & \scriptsize$\pm$0.05 &\scriptsize$\pm$0.06 \\

\multirow{2}{*}{(d)} & 74.12 &  & 74.36 & 74.58 & 74.29 & 74.30 &74.47 & 74.55 &  \textbf{74.78} \\
& \scriptsize$\pm$0.20 &  & \scriptsize$\pm$0.07 & \scriptsize$\pm$0.05 & \scriptsize$\pm$0.24 &\scriptsize$\pm$0.12 &\scriptsize$\pm$0.07  & \scriptsize$\pm$0.09 &\scriptsize$\pm$0.07\\

\bottomrule

\end{tabular} }
\end{sc}
\end{small}
\end{center}

\vskip -0.1in

\label{table:results_CINIC10}
\end{table*}

\section{Experiments} \label{sec:exp}

\subsection{Dataset and Experimental Settings}
\textbf{Datasets.} The CIFAR-10 \cite{Krizhevsky_2009_17719} dataset consists of 60k images distributed among 10 classes, with each class including 5k and 1k images for training and testing, respectively. Each image is a 32$\times$32 sized RGB image. The experiments on CIFAR-10 allows us to evaluate our model's efficacy with less time consumption. We extend our experiments on CINIC-10 \cite{darlow2018cinic} that augments CIFAR-10 formatting but includes a larger set of 270k images whose scale closer to ImageNet. The images are evenly split into each `train', `validate', and `test' sets, with ten classes of 9k images per class. The size of the images is 32$\times$32 as well.

\textbf{Experimental Settings.}
In generating PIs by TDA, by referring to the previous study \cite{som2020pi}, we set birth-time range and Gaussian function parameter as $[$0, 0.3$]$ and 0.01. The threshold for life-time is set to 0.02. we set $g$ and $c$ of PI as 50 and 6, respectively.

For experiments, we set the batch size $b$ as 128, the total epochs as 200 using SGD with momentum 0.9, and a weight decay of $1\times10^{-4}$. The initial learning rate $lr$ is set to 0.1 that is decayed by a factor of 0.2 at epochs 40, 80, 120, and 160. Empirically, we set KD hyperparameters $\lambda$, $\tau$, and $\gamma$ as ($\lambda$ = 0.9, $\tau$ = 4, $\gamma$ = 3000) and ($\lambda$ = 0.6, $\tau$ = 16, $\gamma$ = 2000) for CIFAR-10 and CINIC-10, respectively, referred to previous studies \cite{cho2019efficacy, tung2019similarity, jeon2023leveraging}.

We compare with KD based baselines including conventional KD \cite{hinton2015distilling}, attention transfer (AT) \cite{zagoruyko2016paying}, relational knowledge distillation (RKD) \cite{park2019relational}, variational information distillation (VID) \cite{ahn2019variational}, similarity-preserving knowledge distillation (SP) \cite{tung2019similarity}, attentive feature distillation and selection (AFDS) \cite{wang2019pay}, and multi-teacher based distillation using topological features in KD (Base) \cite{ejasilomar}. AT and SP are utilized with KD. For all baseline methods, the same hyperparameter settings are used as those specified in their papers, and their author-provided code is used for evaluation. 
$\alpha$ of Base is 0.9, and TGD is 0.99 as a default setting. All experiments were repeated three times, and the averaged accuracy and the standard deviation of performance are reported. More details are explained in appendix.

\subsection{Analysis on Teacher-Student Combinations}
In this section, we show analysis on various combinations including different capacity of teachers and architectural styles of teacher-student networks.
\subsubsection{Effect of Teacher Capacity}
We explore the performance of various methods on different types of teacher-student combinations, where the teachers have different capacity. Note, Teacher1 and Teacher2 denote models learned with the raw image data and PI, respectively, and Student denotes a model trained from scratch. As explained in Table \ref{table:info_settings}, we set four different setups for combinations which consist of same or different structures.
We utilize Wide-ResNet (WRN) \cite{zagoruyko2016wide} to construct various compression types of teachers and a student.

As explained in Table \ref{table:results_CIFAR10}, in most of cases, TGD outperforms baselines. Base is an approach using topological features in KD. Compared to KD, Base achieves better performance. For TGD, compared to setup (d), (a) and (b) show better performance, which implies that when teachers have similar architectures to the student, a better student can be distilled. For setup (b), TGD distills a student which is even better than its teachers. Furthermore, (b) of TGD shows better results than (a) and (d) even if their teachers are larger and better than teachers of (b).

In Table \ref{table:results_CINIC10}, TGD shows the best in all cases. Setup (b) of TGD achives better performance than (a) and (d) cases, which implying that a larger or better teacher does not always generate a superior student, as studied in prior works \cite{cho2019efficacy}. Also, if channel of networks for teachers and student is similar, a better student can be distilled compared to other combinations.
We discuss about Teacher2 in Section \ref{pi_analysis}.


\subsubsection{Different Combinations of Teachers and Student}
To analyze the performance with more diverse combinations of teacher-student, we conduct experiments using heterogeneous architectures. Also, we construct teachers with different depth or channel of networks to investigate the interaction and effects between the two teachers.

\textbf{Heterogeneous Architectures of Teacher-Student.}
To explore the effectiveness on heterogeneous teachers and students combinations, we construct combiations with different architectures using WRN \cite{zagoruyko2016wide}, ResNet \cite{he2016deep}, and MobileNetV2 (M.NetV2) \cite{sandler2018mobilenetv2}. We applied the same settings as in the experiments of the previous section.

\begin{table}[htb!]
\centering
\vskip -0.1in
\renewcommand{\tabcolsep}{1.2mm} 
\caption{Accuracy ($\%$) with various knowledge distillation methods for different structure of teachers and students on CIFAR-10. Numbers in brackets denote the number of trainable parameters and accuracy for classification task.}

\begin{center}
\begin{small}
\begin{sc}
\scalebox{0.9}{
\begin{tabular}{c |c c| c c |c }

\toprule
\centering

\multirow{4}{*}{Teacher1} & WRN & WRN & \multirow{2}{*}{vgg13} & WRN & M.Net \\
& 28-1 & 16-8 & & 16-3 & V2 \\
& (0.4M, & (11.0M, & (9.4M, & (1.5M, & (0.6M,  \\
& 85.84) & 89.50) & 88.56) & 88.15) & 89.61)  \\

\midrule
\multirow{3}{*}{Student} & \multicolumn{2}{c|}{vgg8} & \multicolumn{2}{c|}{ResNet20} & WRN28-1  \\
 & \multicolumn{2}{c|}{(3.9M,} & \multicolumn{2}{c|}{(0.3M,} & (0.4M, \\
 & \multicolumn{2}{c|}{85.35{\scriptsize$\pm$0.07})} & \multicolumn{2}{c|}{85.08{\scriptsize$\pm$0.13})} & 85.73{\scriptsize$\pm$0.06}) \\
\midrule 
\multirow{2}{*}{KD} & 86.79 & 86.59 & 85.17 & 85.69 & 87.86\\
 & {\scriptsize$\pm$0.04} & {\scriptsize$\pm$0.17} & {\scriptsize$\pm$0.04}& {\scriptsize$\pm$0.10} & {\scriptsize$\pm$0.15} \\
\multirow{2}{*}{AT} & 87.05 & 87.18 & 85.45 & 86.31 & 88.80 \\
 & {\scriptsize$\pm$0.11} & {\scriptsize$\pm$0.12} & {\scriptsize$\pm$0.32}& {\scriptsize$\pm$0.15}& {\scriptsize$\pm$0.09}\\
\multirow{2}{*}{SP} & 87.11 & 86.73 & 84.94 & 86.33 & 88.84 \\
 & {\scriptsize$\pm$0.22} & {\scriptsize$\pm$0.06} & {\scriptsize$\pm$0.05}& {\scriptsize$\pm$0.09}& {\scriptsize$\pm$0.10} \\

\multirow{2}{*}{Base} & 86.97 & 86.93 & 85.56 & 86.32 & 88.05\\
& {\scriptsize$\pm$0.07} & {\scriptsize$\pm$0.09} & {\scriptsize$\pm$0.20} & {\scriptsize$\pm$0.20}& {\scriptsize$\pm$0.11} \\

\midrule
\multirow{2}{*}{TGD} & \textbf{88.03} & \textbf{87.28} & \textbf{85.64} & \textbf{86.48} & \textbf{88.91} \\
& {\scriptsize$\pm$0.07} & {\scriptsize$\pm$0.04} & {\scriptsize$\pm$0.13} & {\scriptsize$\pm$0.05}& {\scriptsize$\pm$0.08} \\
\bottomrule

\end{tabular}
}
\end{sc}
\end{small}
\end{center}

\vskip -0.1in
\label{table:heteroT}
\end{table}

\begin{table}[htb!]
\centering
\renewcommand{\tabcolsep}{1.2mm} 
\caption{Accuracy ($\%$) with various knowledge distillation methods for different structure of teachers and students on CINIC-10.}

\begin{center}
\begin{small}
\begin{sc}
\scalebox{0.9}{
\begin{tabular}{c |c c c c c |c }

\toprule
\centering

(Teacher1, & \multirow{2}{*}{Student} & \multirow{2}{*}{KD} & \multirow{2}{*}{AT} & \multirow{2}{*}{SP} & \multirow{2}{*}{Base} & \multirow{2}{*}{TGD} \\
Student) & & & & & \\

\midrule
\multirow{2}{*}{\makecell{(WRN16-3,\\ResNet20)}}  & \multirow{4}{*}{\makecell{72.64\\{\scriptsize$\pm$0.13}}} & 74.99 & 74.95 & 75.16 & 74.61 & \textbf{75.28}\\
 & & {\scriptsize$\pm$0.11} & {\scriptsize$\pm$0.19} & {\scriptsize$\pm$0.05}& {\scriptsize$\pm$0.04} & {\scriptsize$\pm$0.03} \\
\multirow{2}{*}{\makecell{(WRN28-3,\\ResNet20)}} & & 74.89 & 75.05 & 75.39 & 74.90 & \textbf{75.47}\\
 & & {\scriptsize$\pm$0.07} & {\scriptsize$\pm$0.15} & {\scriptsize$\pm$0.10}& {\scriptsize$\pm$0.15} & {\scriptsize$\pm$0.03} \\ \bottomrule 

\end{tabular}
}
\end{sc}
\end{small}
\end{center}

\label{table:hetero2T}
\end{table}


In Table \ref{table:heteroT}, TGD outperforms baselines on CIFAR-10. For WRN28-1 teachers and vgg8 student case, the distilled student shows even better performance than its teacher. For vgg13 teachers and ResNet20 student, SP shows even worse than a model learned from scratch. Compared to baselines, TGD achieves better performance, implying topological features help improving performance in KD. In Table \ref{table:hetero2T}, TGD performs better than baselines on CINIC-10 and similar tendency of results on CIFAR-10.
These results also corroborate that better teacher does not guarantee to generate better student \cite{cho2019efficacy}.

\begin{figure}[htb!]
\vskip -0.05in
\begin{center}
\includegraphics[width = 0.35\textwidth]{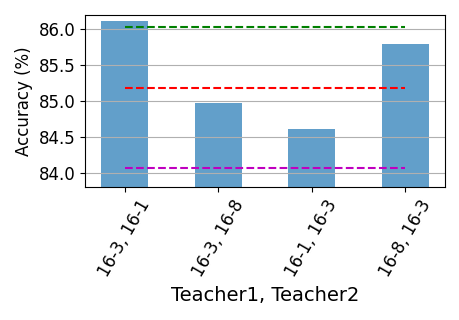}
\centering
\vskip -0.2in 
\caption{Accuracy $(\%)$ of students (WRN16-1) distilled by TGD with various combinations of teachers on CIFAR-10. Teacher1 and Teacher2 consist of different (depth)-(channel) of WRN. Green, red, and magenta dashed lines denote TGD (16-3, 16-3), KD (16-3 Teacher1), and Student (WRN16-1), respectively.}
\label{figure:diffT_cifar10}
\end{center}
\vskip -0.1in 
\end{figure}

\textbf{Analysis on Different Teachers.}
To investigate the effect of each teacher on distillation, we construct Teacher1 and Teacher2 with different depth or channel of WRN. As shown in Figure \ref{figure:diffT_cifar10}, when the network capacity of Teacher2 is smaller than that of Teacher1, a better student is distilled, which shows that topological features act as complementary features to those from the raw image data. For (16-3, 16-1) and (16-8, 16-3) cases, (16-3, 16-1) shows better results and even better than (16-3, 16-3). This presents that (16-3, 16-1) case generates knowledge that is well matched with the student and stronger than the one from other combinations, which alleviates performance degradation issues that may arise due to knowledge gaps.


\subsection{Ablations and Sensitivity Analysis}
In this section, we investigate sensitivity for $\alpha$, robustness on noise, and evaluate with feature visualization and model reliability.

\subsubsection{Effect of $\alpha$ hyperparameter}


\begin{figure}[htb!]
\begin{center}
\includegraphics[width = 0.48\textwidth]{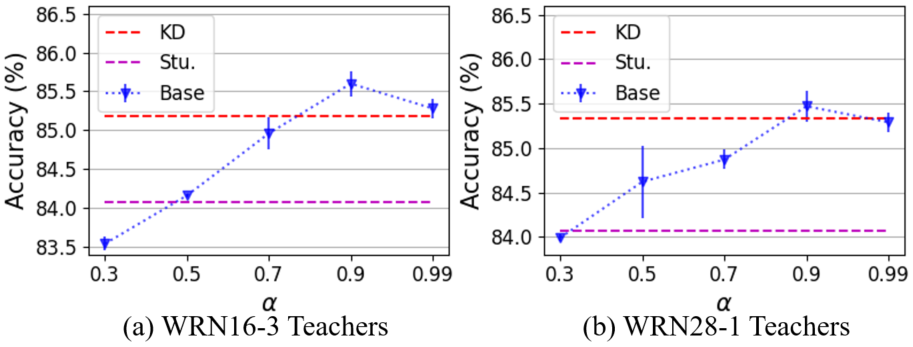}
\centering
\vskip -0.1in
\caption{Accuracy ($\%$) of students (WRN16-1) for various methods with different $\alpha$ on CIFAR-10.}
\label{figure:alpha_cifar10}
\end{center}
\vskip -0.05in
\end{figure}

We investigate performance of Base on various $\alpha$ hyperparameters. Note, Stu. denotes student trained from scratch. As illustrated in Figure \ref{figure:alpha_cifar10}, over 0.7 shows better results than KD using a single teacher trained with the raw image data. This implies that relying on Teacher1 more than Teacher2 is effective in distilling a superior student. This tendency is also the same on CINIC-10, which is different from using topological features on time-series data: their optimal $\alpha$ is vary across datasets (e.g. 0.7 or 0.3) \cite{ejasilomar, jeon2024topological}.
The fact that a high $\alpha$ indicates good results implies that Teacher1 provides stronger information than Teacher2, which is well matched with the student model, since Teacher1 and the student are trained with the same representations and possess similar statistical characteristics. However, using excessive $\alpha$ does not provide the best, which implies topological features indeed act as complement features in distillation to improve performance.
With this observations on Base, we utilized high $\alpha$ values which are larger than 0.7. For experiments of previous section, TGD uses 0.99 which is high $\alpha$ and shows the best results. This is because using an annealing strategy encourages a student to preserve features of the raw image data, which are better matched with Teacher1. By leveraging topological features, TGD outperforms baselines including Base. 
More results are described in appendix.

\subsubsection{Analysis on Persistence Image} \label{pi_analysis}

\begin{figure}[htb!]
\vskip -0.05in
\begin{center}
\includegraphics[width = 0.39\textwidth]{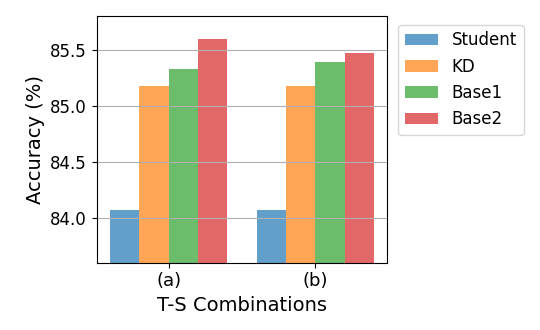}
\centering
\vskip -0.2in
\caption{Accuracy ($\%$) of students (WRN16-1) for various methods with setup (a) and (b) on CIFAR-10.}
\label{figure:pi_cifar10}
\end{center}
\vskip -0.1in

\end{figure}

We use sublevel-set filtration to create PI from an image through TDA, which is simpler than other methods but useful in topological feature extraction \cite{barnes2021comparative}. As explained in \cite{barnes2021comparative}, coordinate transforms can affect to extracting topological features. To collect diverse and richer features, multi-scale or multiple coordinate transforms can be leveraged. In our experiments, we used row- and column-wise transforms which collect topological features differently and generate 6 channels of PI.
Since datasets in our experiments have complicated patterns (e.g. complex background and multiple channels with diverse region or size of targets), using PI solely cannot show good results. In most cases, performance was close to 35$\%$ and 33$\%$ in terms of classification accuracy for CIFAR-10 and CINIC-10, respectively.
The performance of this model itself is not very good, but when it is included in KD process, it has the advantage of providing useful information that complements features from the original data and helps improving performance, which can be observed in empirical evaluations. Note, Base1 denotes using row-wise transform to generate 3 channels ($P_{Rr}$, $P_{Gr}$, $P_{Br}$) of PIs, and Base2 denotes utilizing multi-wise transforms to create 6 channels of PIs. As shown in Figure \ref{figure:pi_cifar10}, Base1 outperforms conventional KD that uses a single teacher trained with the original image data. Using multi-wise transforms, Base2, helps distilling a better student. Thus, providing more diverse topological information can generate a superior student. 
Also, these results represent the compatibility of topological features in distillation for performance improvement.

\subsubsection{Robustness to Noise}
Topological features have shown an excellent ability to withstand noise and perturbations. To explore this, we evaluate student models on a different level of noises for testing data. To inject noises, we utilize Gaussian noise with different levels. Specifically, we apply randomly chosen Gaussian kernel standard deviation from 0.01 to the selected parameter of $\sigma$. The kernel size is set as 5$\times$5. The levels of noises are defined by $\sigma$ as follows; Level 1 (0.5), Level 2 (1.0), Level 3 (1.5), Level 4 (2.0).

\begin{figure}[htb!]
\vskip -0.1in
\begin{center}
\includegraphics[width = 0.34\textwidth]{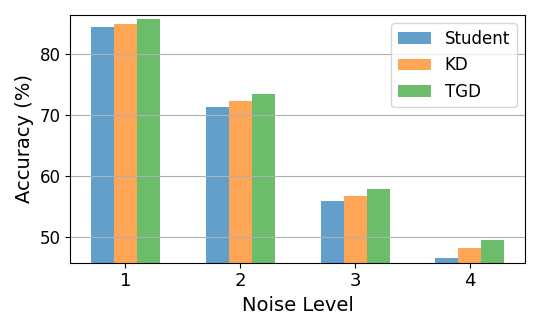}
\centering
\vskip -0.15in
\caption{Accuracy ($\%$) of students (WRN16-1) for various corruption severity levels on CIFAR-10. WRN16-3 teachers are utilized.}
\label{figure:noise_cifar10}
\end{center}
\vskip -0.15in
\end{figure}

As shown in Figure \ref{figure:noise_cifar10}, as the level of noise increases, the performance of baselines deteriorates significantly, but TGD can withstand the noise much better. This represents that topological features aid in distilling a superior student to withstand noise.

\subsubsection{Visualization of Models}
To study the behavior of models and characteristics of extracted features intuitively, we visualize features with diverse methods such as similarity maps and activation maps.

\begin{figure}[htb!]
\begin{center}
\includegraphics[width = 0.42\textwidth]{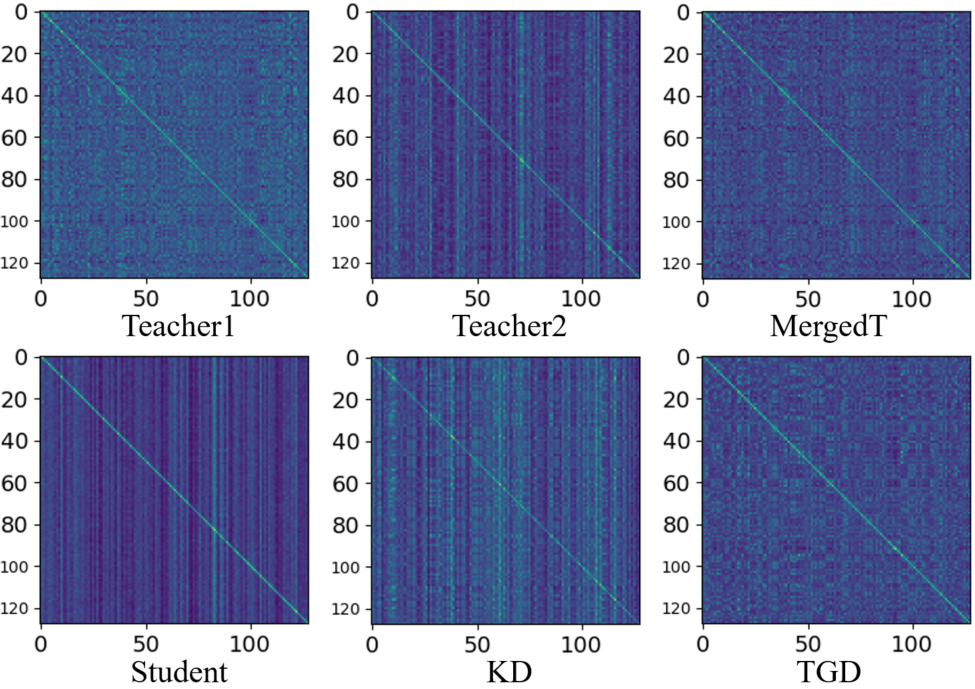}
\centering
\vskip -0.1in
\caption{An illustration of similarities for various methods on CIFAR-10. WRN16-3 teachers and WRN16-1 student are utilized.}
\label{figure:highlevel_sim}
\end{center}
\vskip -0.2in
\end{figure}

\textbf{Analysis of Feature Map.}
To explore similarity maps of different methods, we visualize the similarities of high-level intermediate layers that provides more distinguishable maps between methods intuitively, as shown in Figure \ref{figure:highlevel_sim}. Student models distilled from diverse methods are used for visualization. MergedT denotes the similarities of integrated features from two teachers, which includes topological features. Student and KD present similar patterns showing column-wise contrasts, which rely on image data alone. TGD shows block-wise patterns that are similar to MergedT, which differs from Student and KD. This represents that TGD encourages a student to obtain topological features, which enables to obtain improved performance. More details are provided in appendix.

\begin{figure}[htb!]
\begin{center}
\includegraphics[width = 0.38\textwidth]{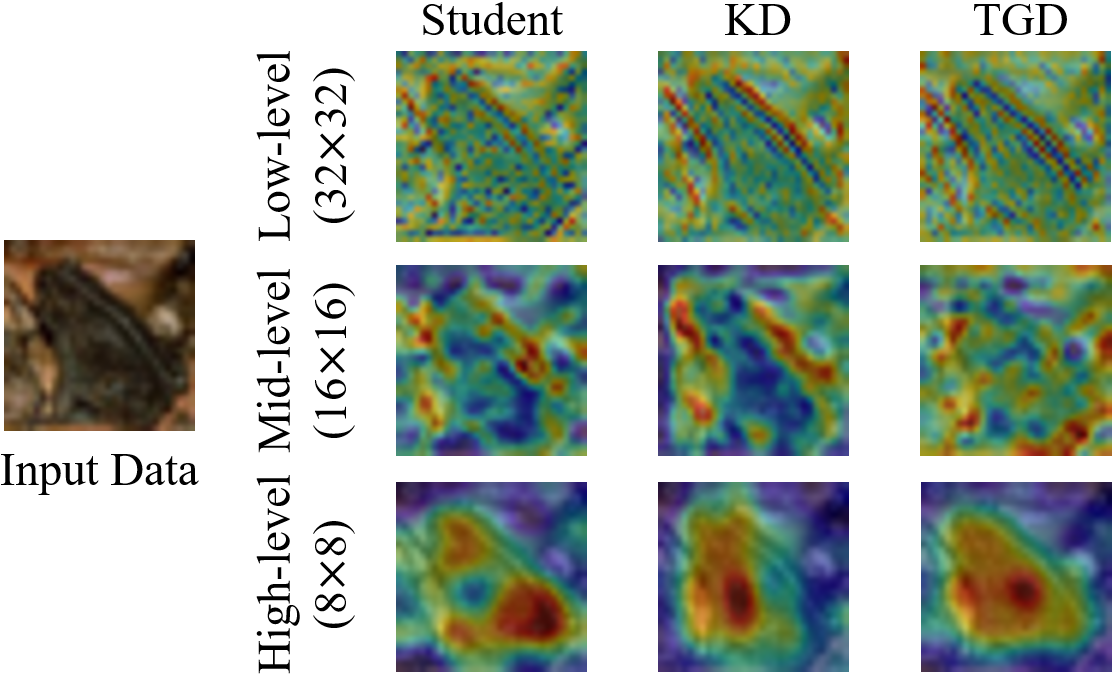}
\centering
\vskip -0.1in
\caption{Activation maps for various methods, on a frog image.}
\label{figure:gradcam_lv}
\end{center}
\end{figure}

\begin{figure}[htb!]
\begin{center}
\includegraphics[width = 0.48\textwidth]{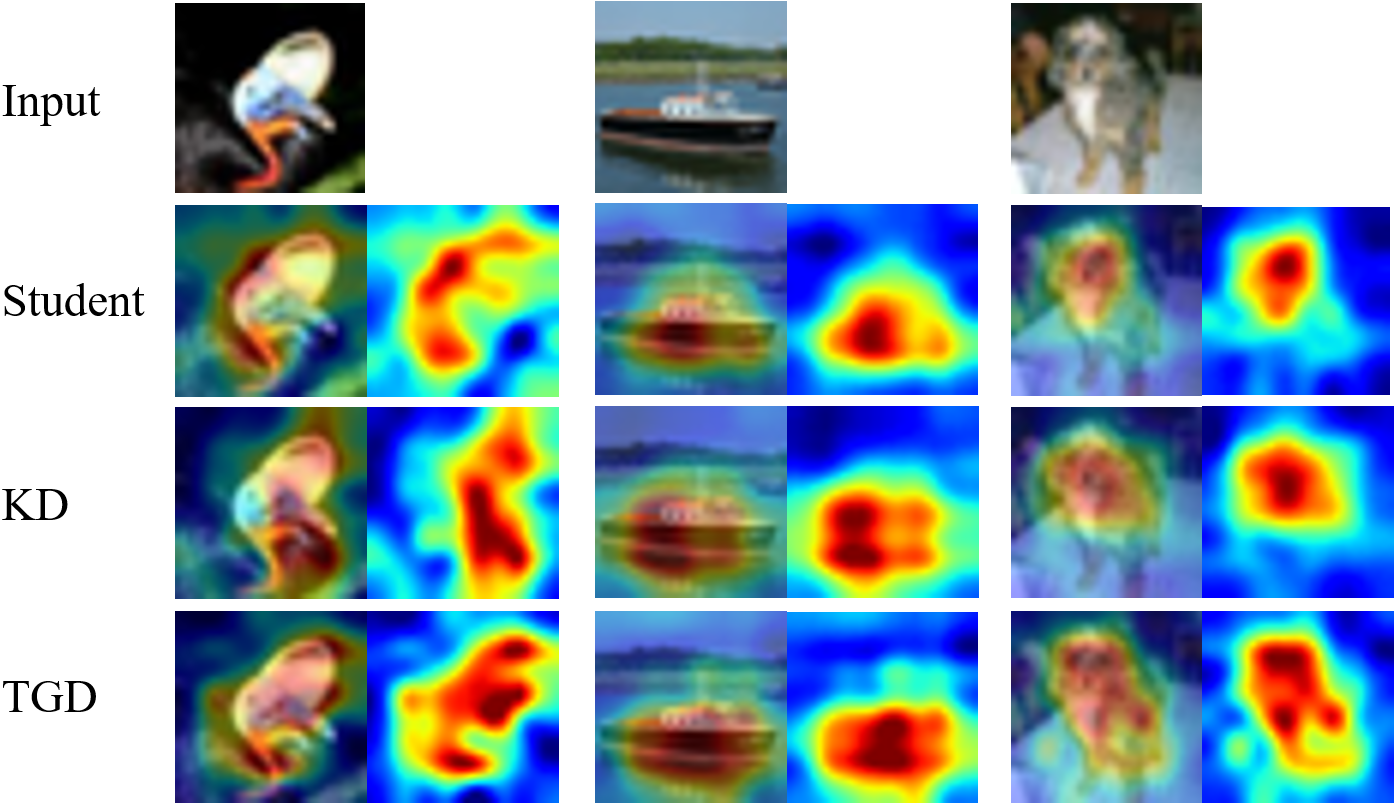}
\centering
\vskip -0.1in
\caption{Activation maps of high-level layer for various methods on bird, ship, and dog images.}
\label{figure:gradcam_cls}
\end{center}
\vskip -0.1in
\end{figure}

\textbf{Analysis of Activation Map.}
We visualize the activation maps of various methods by Grad-CAM \cite{selvaraju2017grad} to analyze the coarse localization map of the important regions of each model with various intermediate layers. WRN16-3 teachers and WRN16-1 student are utilized for activation map visualization. In Figure \ref{figure:gradcam_lv}, each method focuses on different locations across different intermediate layers. Compared to other methods, TGD focuses on whole area of a target object, which is recognizable intuitively in maps from the high-level layer.
We also visualize maps of high-level layer on different input data, as shown in Figure \ref{figure:gradcam_cls}. Compared to other methods, TGD distinctly focuses more on the target area with high weight and less on background regions, which indicates that TGD has better classification ability. More results are provided in appendix.

\subsubsection{Model Reliability}


To investigate the generalizability of models, we computed expected calibration error (ECE) \cite{guo2017calibration} and negative log likelihood (NLL) \cite{guo2017calibration}. ECE is to measure calibration errors, implying the reliability of a model. NLL represents the probabilistic quality of the model.

As explained in Table \ref{table:reliability_result}, for both setups, TGD shows lower ECE and NLL compared to baselines, which implies topological features aid in improving not only for accuracy but also for generalizability. 

\begin{table}[htb!]
\vskip -0.1in
\centering
\caption{ECE ($\%$) and NLL ($\%$) for various knowledge distillation methods on CIFAR-10. The results (ECE, NLL) for WRN16-3 and WRN28-1 teachers (Teacher1) are (1.469$\%$, 44.42$\%$) and (2.108$\%$, 64.38$\%$), respectively. (2.273$\%$, 70.49 $\%$) for WRN16-1 Student.}
\vskip 0.05in
\begin{center}
\begin{small}
\begin{sc}

 \scalebox{0.92}{
\begin{tabular}{c |c c | c c}

\toprule
\centering

\multirow{2}{*}{Method} & \multicolumn{2}{c|}{Setup (a)}&  \multicolumn{2}{c}{Setup (b)}\\ \cmidrule{2-5} 
& ECE & NLL & ECE & NLL \\
\midrule

KD  & 2.035 & 62.26 & 2.188  & 67.21  \\

AT & 1.978 & 60.48 & 2.156 & 67.14  \\

TGD & \textbf{1.865} & \textbf{56.05} & \textbf{1.940}  & \textbf{60.12}  \\

\bottomrule

\end{tabular}
 }


\end{sc}
\end{small}
\end{center}

\vskip -0.05in
\label{table:reliability_result}
\end{table}

\subsection{Processing Time}
We measure the processing time of various models on CIFAR-10 testing set (10k samples). The total processing time is explained in Table \ref{table:processing_time}. A student (WRN16-1) of TGD takes much less time than teachers on both CPU and GPU. Creating PI (6 channels) takes more than 30k seconds on the CPU, which is not efficient in inference time as well. As described in the prior section, the student by TGD outperforms a model learned from scratch by 1.96$\%$ in classification accuracy. These findings clearly highlight the essential necessity of using a compact model for implementation on small devices with limited computational resources and the effectiveness of TGD.

\begin{table}[ht!]
\renewcommand{\tabcolsep}{1.2mm} 

\vskip -0.1in
\caption{Processing time on CIFAR-10 testing set.}
\label{table:processing_time}

\vskip 0.05in

\begin{center}
\begin{small}
\begin{sc}

\begin{tabular}{c |c c c  c}
\toprule
\multirow{2}{*}{Method} & Teacher1 & Teacher2 & \multicolumn{2}{c}{TGD} \\

 \cmidrule{2-5}
 & WRN16-3 & WRN16-3 & \multicolumn{2}{c}{WRN16-1} \\
\midrule

\multirow{2}{*}{GPU (sec)} & \multirow{2}{*}{67.83} & 10280 (PI on CPU) & \multicolumn{2}{c}{\multirow{2}{*}{\textbf{60.81}}} \\
 &  & +90.11 (model) & \multicolumn{2}{c}{}\\ \midrule
\multirow{2}{*}{CPU (sec)} & \multirow{2}{*}{263.31} & 10280 (PI on CPU) & \multicolumn{2}{c}{\multirow{2}{*}{\textbf{90.20}}} \\ 
 & & +449.48 (model) & \multicolumn{2}{c}{}\\
\bottomrule
\end{tabular}

\end{sc}
\end{small}
\end{center}
\vskip -0.1in
\end{table}

\section{Discussion}
Based on the empirical results, we explored the effectiveness of TGD with various combinations of teachers and students. Also, we investigated characteristics of model behaviors by visualization of similarity maps from intermediate layers.


The focus of this paper is to leverage multiple teachers in KD for transferring topological features to a student, which is to obtain a small-sized and superior model. Utilizing multiple teachers can increase the computational cost in KD training process, however a single distilled model from our approach, TGD, has the advantage of not requiring additional data or layers at test-time after learning once. Also, TGD does not include hidden layers in knowledge transfer process, which does not require much computational cost for fusing different features to utilize multiple teachers. Recently, methods such as the teacher selection strategy \cite{shang2023multi} have been studied to save resources during training time. Reducing computing resources by using multiple teachers trained with different representations requires further exploration.

Teacher2 models trained from scratch with PI only show much worse accuracy in classification tasks compared to Teacher1 and Student models. The performance of Teacher2 is explained in more detail in appendix. However, the network model of Teacher2 is utilized as a teacher in KD to create a superior student by synergizing with Teacher1. The student possession of topological features is observed in similarity maps from intermediate layers.
Not only are the transforms in filtration considered in this paper, but there are also various methods to create PI.
Specifically, we used simplicial homology in this paper, which is the standard approach in many applications \cite{barnes2021comparative}. The other popular methods such as cubical homology \cite{Kaczynski2004} and multiple density areas \cite{barnes2021comparative} can be utilized to extract useful features for the same purpose.
Additionally, there is still much room for improving performance with a more advanced Teacher2, which can be analyzed with empirical experimentation. This can be more explored in a future work.

\section{Conclusion}
In this paper, we present a framework for leveraging topological features in KD with multiple teachers and feature similarities for image data analysis. We demonstrated the effectiveness of utilizing topological features in KD based on the proposed method, TGD, under various evaluations, including different combinations of teachers and students, feature and activation maps visualization, and resistance to noise, with empirical testing on classification task.

In future work, more advanced ways to compute persistence features, including transform-based approaches, can be explored in improving performance, such as using cubical persistent homology \cite{Kaczynski2004} in filtration. Also, more challenging test-conditions can be explored to highlight where TDA features provide robustness in the context of computer vision applications.

\section*{Acknowledgements}
This work was supported by NSF grant 2323086.

\nocite{langley00}

\bibliography{main_camera_ready}
\bibliographystyle{icml2024}

\newpage
\appendix
\onecolumn

\begin{appendices}
We provide additional experimental settings and results. Also, more details and our findings are discussed.
For reproducibility, the source codes, models, etc., are available at \url{https://github.com/jeunsom/TGD}.

\section{Additional Experiments}
\subsection{Experimental Settings}
For $\lambda$ and $\tau$, we referred to previous studies \cite{cho2019efficacy, tung2019similarity} to choose the popular parameters in KD \cite{cho2019efficacy, tung2019similarity, jeon2023leveraging}.

Since our method uses similarity maps which can be obtained from outputs of intermediate layers, additional techniques including more hidden layers or interpolations are not used. Also, no augmentation method is applied for CIFAR-10 and CINIC-10. 

Life-time threshold denotes points in PD are discarded if their values are less than the threshold.

The all experiments were executed on a desktop equipped with a 2.00 GHz CPU (Intel® Xeon(R) CPU E5-26200 0), 16 GB of memory, and an NVIDIA GeForce GTX 980 graphic card (2048 NVIDIA® CUDA® cores and 4 GB of memory).


\begin{wrapfigure}{R}{7.8cm}
\begin{center}
\vspace{-1.0em}
\includegraphics[width=7.8cm]
{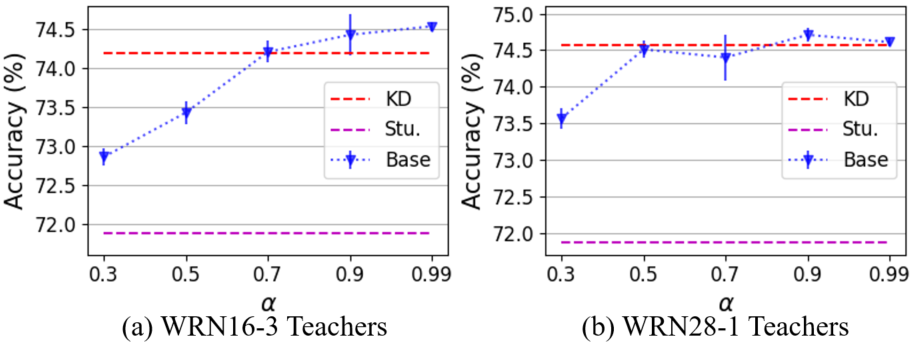}
\centering
\vskip -0.15in
\caption{Accuracy ($\%$) of students (WRN16-1) for various methods on CINIC-10.}
\label{figure:alpha_cinic10}
\end{center}
\vskip -0.2in
\end{wrapfigure}

\subsection{Effect of $\alpha$ Hyperparameter}
In Figure \ref{figure:alpha_cinic10}, results of different methods with various $\alpha$ on CINIC-10 are illustrated. When $\alpha$ is larger than 0.7, the distilled student outperforms baselines. This implies higher weights on Teacher1 generate a superior student. This may because Teacher1's statistical characteristics are more matched with the student, where two models are trained on the same representations of data. Also, this results show that topological features are not stronger but indeed act as complement features to improve the performance. For TGD, 0.99 $\alpha$ shows the best in most of cases since an annealing strategy encourages a student to preserve statistical characteristics of features on the raw image.

\subsection{Visualization of Models}

\begin{figure}[htb!]
\begin{center}
\includegraphics[width = 0.8\textwidth]{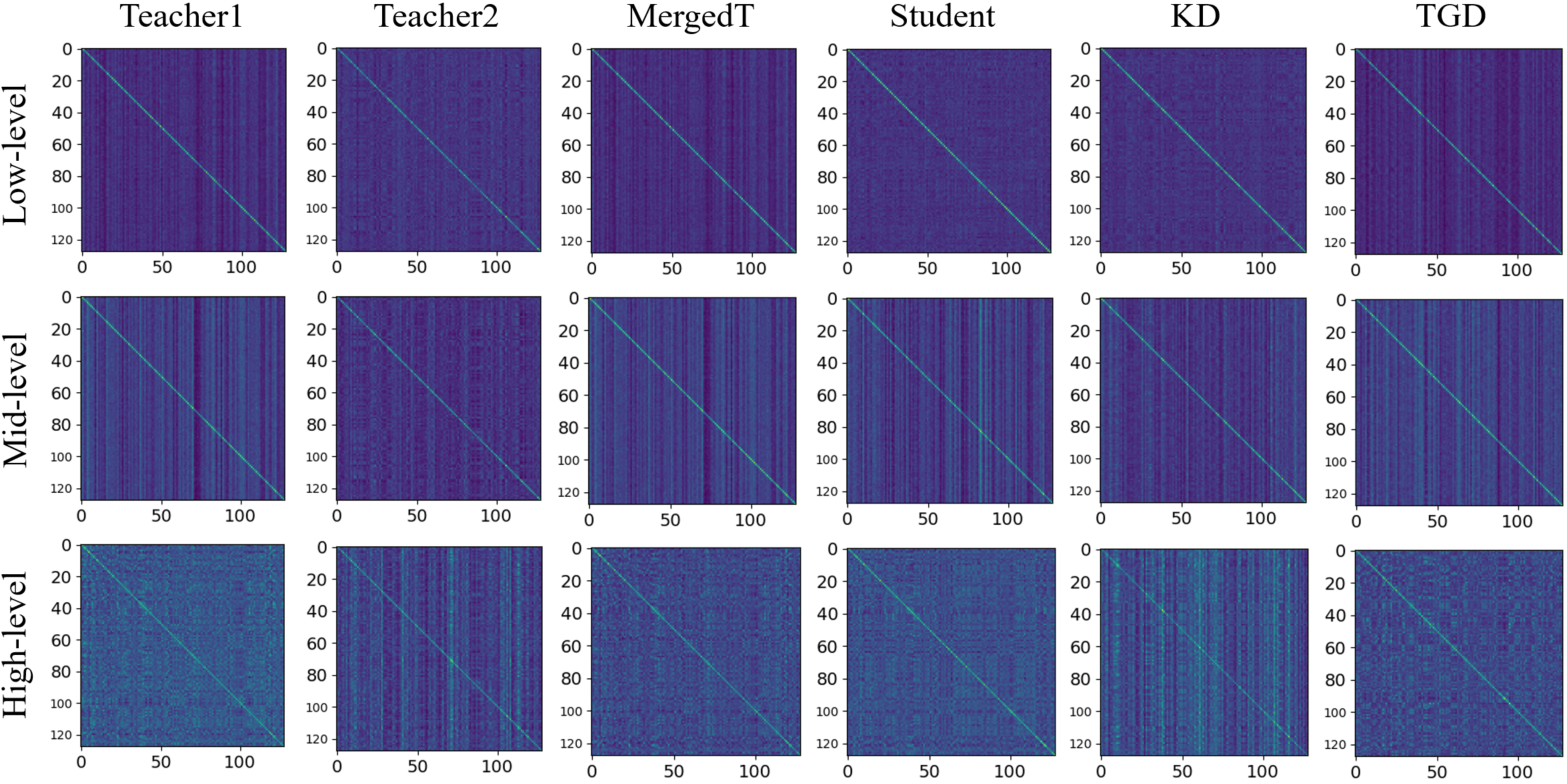}
\centering
\vskip -0.1in
\caption{An illustration of similarities for various methods on CIFAR-10. WRN16-3 teachers and WRN16-1 student are utilized.}
\label{figure:alllevel_sim}
\end{center}
\vskip -0.1in
\end{figure}

\begin{wrapfigure}{R}{7.8cm}
\begin{center}
\includegraphics[width=7.8cm]{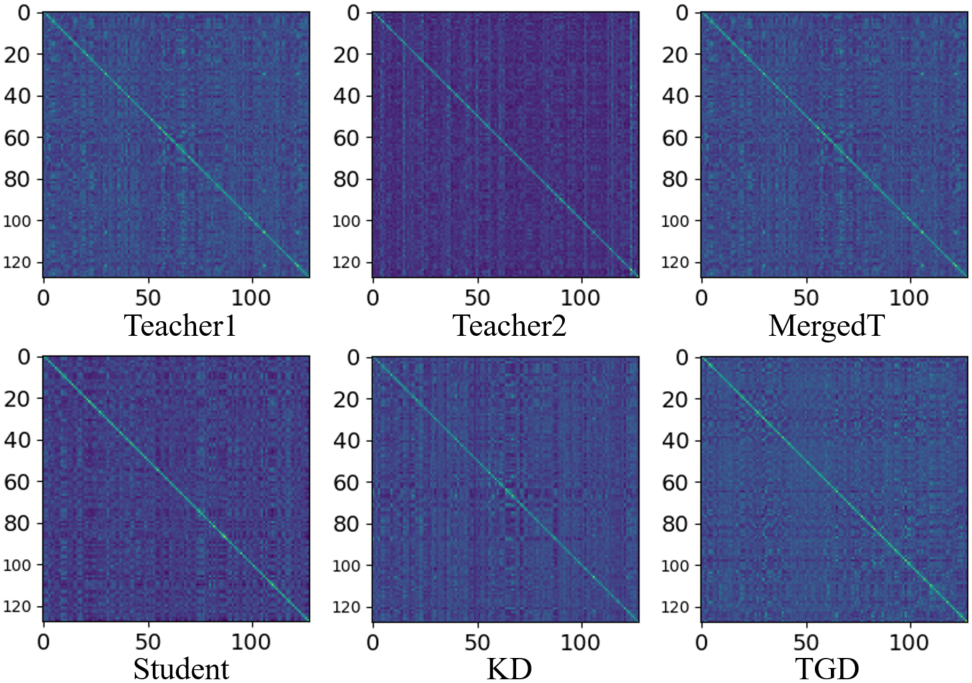}
\centering
\vskip -0.1in
\caption{An illustration of similarities for various methods on CINIC-10. WRN16-3 teachers and WRN16-1 student are utilized.}
\label{figure:highlevel_sim_cinic10}
\end{center}
\vskip -0.2in
\end{wrapfigure}

\textbf{Analysis of Feature Map.}
More results from various intermediate layers are illustrated in Figure \ref{figure:alllevel_sim}. Compared to low-level, similarities of high-level shows more highlighted patterns and more dissimilar characteristics are shown between different methods. Since Teacher1 and Student are trained from scratch with the image data, they possess similar characteristics. However, KD and Student of high-level have different patterns. This shows the effects of KD. However, TGD differs from KD since TGD is trained with MergedT providing topological features in KD learning process. These results represent that a student distilled by TGD possesses topological features, which is superior than using the raw image data alone in training process.

Additionally, we visualize the similarities of student models on CINIC-10 in Figure \ref{figure:highlevel_sim_cinic10}. Student and KD present contrast patterns compared to TGD. TGD shows brighter patterns on correlated points between different samples. Since TGD is trained with MergedT, their patterns and characteristics are more similar. This implies that a distilled student of TGD produces topological features that complements the features from the raw image data.

\begin{figure}[htb!]
\begin{center}
\includegraphics[width = 0.65\textwidth]{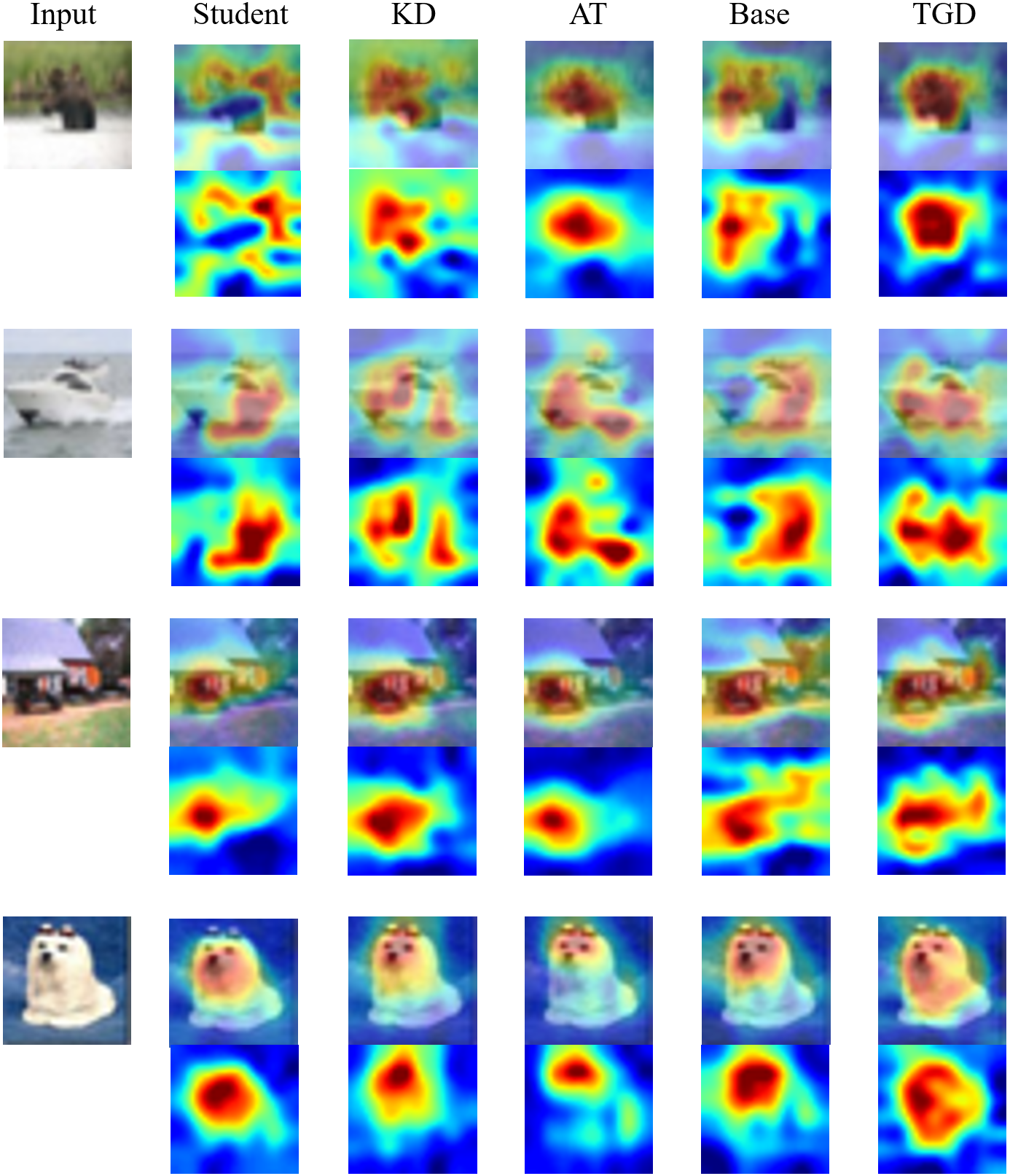}
\centering
\caption{Activation maps of high-level layer for various methods. Labels of input data are deer, ship, truck, and dog.}
\label{figure:gradcam_cls_app}
\end{center}
\end{figure}

\textbf{Analysis of Activation Map.} We provide more activation maps of high-level intermediate layer on different input data. WRN16-3 teachers and WRN16-1 student are utilized. As illustrated in Figure \ref{figure:gradcam_cls_app}, TGD focuses more on the target area with high weight and less on the background area compared to other methods. This implies that TGD has better discrimination ability between target and background regions, leading to better classification performance. Thus, based on TGD, topological features guide a student to obtain better discrimination ability, improving performance in image analysis.

\subsection{Analysis of PI and Teacher2}

\begin{figure}[htb!]
\begin{center}
\includegraphics[width = 0.65\textwidth]{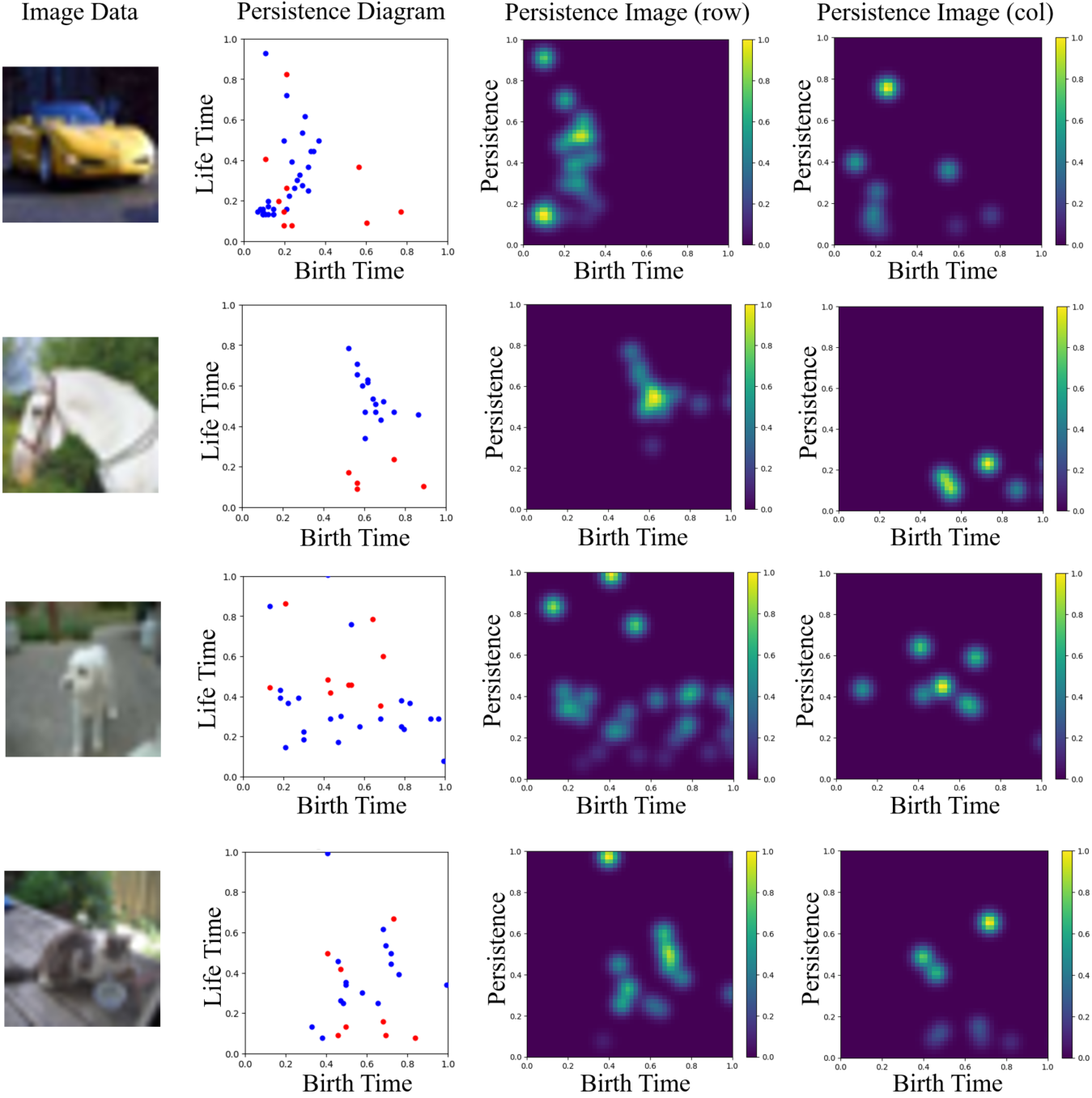}
\centering
\vskip -0.1in
\caption{PD and its corresponding PI. Lifetime points in PD appears bright colors in PI. Red and blue denote points from row- and column-wise transforms, respectively.}
\label{figure:pdpi_all}
\end{center}
\vskip -0.1in
\end{figure}

We illustrate more examples of images and their corresponding PD and PI of ($P_{Rr}$ and $P_{Rc}$) in Figure \ref{figure:pdpi_all}. We visualize the results on different transforms of image for filtration. As shown in the figure, PIs for row- and column-wise transforms are different, which can be observed intuitively. As explained results of Base on leveraging single or multiple transforms, using more diverse topological information is more useful in distillation.

\begin{figure}[htb!]
\begin{center}
\includegraphics[width = 0.65\textwidth]{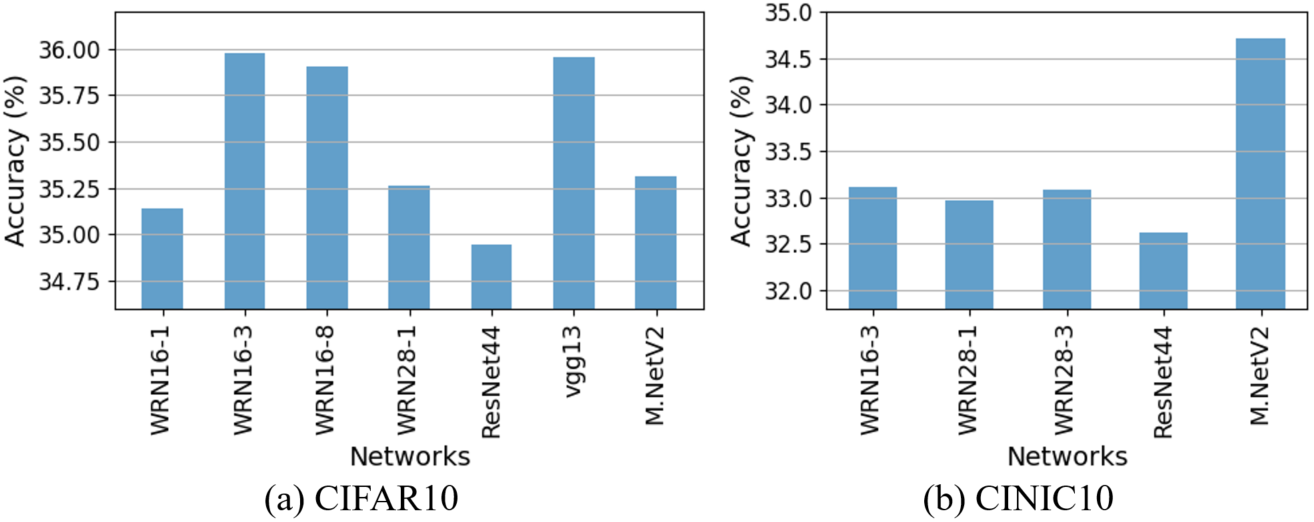}
\centering
\vskip -0.1in
\caption{Accuracy $(\%)$ for various network models trained from scratch with PI.}
\label{figure:pi_model}
\end{center}
\vskip -0.2in
\end{figure}

The models (Teacher2) trained from scratch with PI achieves approximately 35$\%$ and 33$\%$ in overall cases of classification task for CIFAR-10 and CINIC-10, respectively, as shown in Figure \ref{figure:pi_model}.
To train models, 6 channels of PIs are utilized. As explained in the manuscript, using PI solely to train a model does not show good results, which differs from time-series data analysis \cite{ejasilomar, jeon2024topological}. However, this can be combined in KD process and utilized to improve the performance while this provides complementary features, topological features.

\begin{wrapfigure}{R}{8.5cm}
\begin{center}
\vskip -0.7in
\includegraphics[width=5.8cm]{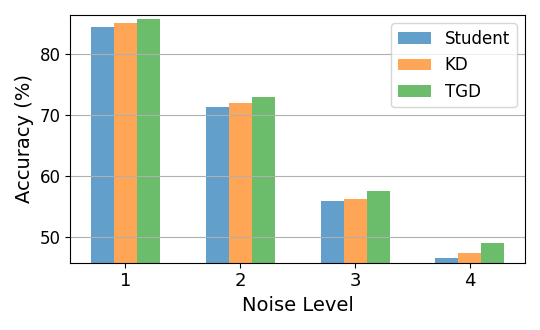}
\centering
\vskip -0.15in
\caption{Accuracy ($\%$) of students (WRN16-1) for various corruption severity levels on CIFAR-10. WRN28-1 teachers are utilized.}
\label{figure:CIFAR10_281noise_Analysis}
\end{center}
\vskip -0.8in
\end{wrapfigure}

\subsection{Robustness to Noise}
We investigate the robustness to noise on students distilled with WRN28-1 teachers by various methods, as illustrated in Figure \ref{figure:CIFAR10_281noise_Analysis}. For noise injection, the settings are the same as explained in the manuscript. In all noise levels, TGD shows the best accuracy. This implies that topological features help the student to obtain better resilience to noise.

\end{appendices}

\end{document}